\documentclass[10pt, journal]{IEEEtran}
\usepackage{graphicx}
\usepackage{cite}
\usepackage{amssymb}
\usepackage[dvipsnames]{xcolor}
\usepackage{amsmath}
\usepackage{float}
\usepackage{hyperref}
\usepackage{algorithm}
\usepackage{algorithmic}
\usepackage{placeins}
\usepackage{bm}
\usepackage{lipsum}

\begin{document}

\title{Deep Reinforcement Learning for Multi-Truck Vehicle Routing Problems with Multi-Leg Demand Routes}

\author{Joshua Levin, Randall Correll, Takanori Ide, Suzuki Takafumi, Saito Takaho, Alan Arai
\thanks{Joshua Levin and Randall Correll are with QC Ware Corp., Palo Alto, CA USA}
\thanks{Takanori Ide is with the Department of Mathematics and Information Science at Josai University, Chiyoda-ku, Tokyo, Japan}
\thanks{Suzuki Takafumi and Saito Takaho, are with AISIN CORPORATION, Tokyo Research Center, Chiyoda-ku, Tokyo, Japan}
\thanks{Alan Arai is with Aisin Technical Center of America, San Jose, CA USA}
}

%\author{Takefumi Kawakami}
%\affiliation{
%AISIN CORPORATION, Tokyo Research Center, Chiyoda-ku, Tokyo, Japan
%}
%\author{Noriyuki Miyake}
%\affiliation{
%AISIN CORPORATION, Tokyo Research Center, Chiyoda-ku, Tokyo, Japan
%}
%\author{Takanori Ide}
%\affiliation{
%AISIN CORPORATION, Tokyo Research %Center, Chiyoda-ku, Tokyo, Japan
%}
%\author{Takafumi Suzuki}
%\affiliation{AISIN CORPORATION, Kariya %city, Aichi, Japan}
%\author{Takaho Saito}
%\affiliation{
%AISIN CORPORATION, Tokyo Research %Center, Chiyoda-ku, Tokyo, Japan
%}
%\author{Alan Arai}
%\affiliation{Aisin Technical Center of %America, San Jose, CA USA}
\maketitle
\begin{abstract}
    Deep reinforcement learning (RL) has been shown to be effective in producing approximate solutions to some vehicle routing problems (VRPs), especially when using policies generated by encoder-decoder attention mechanisms. While these techniques have been quite successful for relatively simple problem instances, there are still under-researched and highly complex VRP variants for which no effective RL method has been demonstrated. In this work we focus on one such VRP variant, which contains multiple trucks and multi-leg routing requirements. In these problems, demand is required to move along sequences of nodes, instead of just from a start node to an end node. With the goal of making deep RL a viable strategy for real-world industrial-scale supply chain logistics, we develop new extensions to existing encoder-decoder attention models which allow them to handle multiple trucks and multi-leg routing requirements. Our models have the advantage that they can be trained for a small number of trucks and nodes, and then embedded into a large supply chain to yield solutions for larger numbers of trucks and nodes. We test our approach on a real supply chain environment arising in the operations of Japanese automotive parts manufacturer Aisin Corporation, and find that our algorithm outperforms Aisin's previous best solution.
\end{abstract}
\

\section{Introduction}
Vehicle Routing Problems (VRPs) \cite{dantzig1959truck, toth2014vehicle, Han_2018} are NP-hard combinatorial optimization problems in which one or more trucks must deliver material between several different locations. VRPs arise frequently in the context of supply chain logistics. Often, the complexity of a vehicle routing problem acts as a computational bottleneck limiting the efficiency of supply chain operations. The nature of such computational bottlenecks is that even a modest improvement in the quality of solutions can lead to significant benefits, both financial and environmental. 

VRPs have been thoroughly studied in operations research, which has led to many different VRP variations. Examples include the capacitated VRP (CVRP) in which the trucks have fixed carrying capacities, the VRP with time windows (VRPTW) in which demand must arrive at its destination within a specified time interval, the split-delivery VRP (SDVRP) in which trucks are allowed to load a subset of the demand at a single location and leave the rest to be picked up later, and many others. But there are some variations of the VRP that arise in real-world supply chains which have not been thoroughly studied. One example is VRPs where the demand has multi-leg routing requirements. In these problems, demand is required to move along sequences of locations, and not just from one location to another. This type of problem arises in the supply chain operations of Aisin Corporation, a Japanese automotive manufacturing company, and is the focus of this work.

Recently, there has been a growing interest in solving combinatorial optimization problems using reinforcement learning (RL) \cite{bello2016neural, NIPS2015_29921001}. More specifically, there has been a plethora of research on solving VRPs using RL approaches \cite{nazari2018reinforcement, Li_2022, kool2018attention, Yu2019, Kalakanti2019, Oxenstierna2019, lu2019learning, zheng2021combining, d2020learning, costa2021, Wu2022, NALEPA2020203, Lin2022, Zhao2021, Peng2020, Liao2019, Joe_Lau_2020, geng2021deep, pan2023deep, arishi2023multi, phiboonbanakit2021hybrid, zhou2023reinforcement, mak2023fair, soroka2023deep} (see \cite{raza2022} for a review of recent advances). Many of these models use attention layers, which has proven to be quite effective for solving many simple VRP variations. The use of RL for such problems is very natural. RL is most useful in very complex decision processes, where a decision's consequences can only be learned through trial-and-error. Many games such as Chess and Go fit this description, and indeed the best known solutions for such games use RL methodology \cite{silver2017mastering, silver2018}. A VRP can be thought of as a one-player game in which the player controls the trucks and decides what routes they drive and what they pickup and dropoff at each stop. In this context, VRPs present Markov decision processes for which the consequences of decisions are best learned through experience. 

However, state-of-the-art RL approaches for VRPs are not yet deployable in commercial settings since they typically only address very simple VRP variations. Past work has almost exclusively focused on the case of a single truck whose task is to simply deliver all demand to one special location. Realistic supply chain environments, like the one we consider here, involve many trucks and much more complicated delivery requirements.

In this work, we take steps towards developing RL models that are capable of obtaining good solutions to a multi-truck VRP with multi-leg routing requirements. As our testing ground, we use a real supply chain environment of Aisin Corporation and find that our method outperforms Aisin's solution for a slightly simplified version of this supply chain. We build upon the method of \cite{kool2018attention} (which was designed for single truck VRPs with simple routing requirements) by developing and incorporating new techniques which allow for multiple trucks and multi-leg routing. Importantly, our approach breaks the problem into many smaller problems of fixed size. This allows us to scale up the size of the problem without having to scale the size of our model, by just solving more sub-problems.

The rest of this paper is organized as follows. In Sec. \ref{VRP}, we introduce the realistic VRP formulation which is the focus of this work. In Sec. \ref{Algorithm}, we describe the complete workflow of our algorithm, treating our neural networks as black boxes. In Sec. \ref{Neural Net}, we describe in detail the architecture of our neural networks. In Sec. \ref{Training}, we outline the methods we use to train our neural networks. Finally, in Sec. \ref{Results}, we present the results of deploying our algorithm on the realistic Aisin test case.

This work is an extension of the work done in \cite{correll2022reinforcement}, and was completed after SW and FS left QC Ware Corp.

\section{Vehicle Routing Problems}\label{VRP}

In this section, to highlight the complexity of the realistic VRP on which we will test our method, we will first review a basic version of the VRP. Then we will introduce a generalized and more complex VRP that models an actual supply chain environment that arises in the daily operations of Aisin Corporation.

\subsection{Basic Vehicle Routing Problem}
First, we review the capacitated vehicle routing problem with split deliveries (SDVRP). An instance of the SDVRP is specified by the following data:
\begin{enumerate}
\item A graph $G$ with $n + 1$ nodes $z_{0},z_{1},\ldots,z_{n}$.
\item An $(n+1)\times (n+1)$ matrix $T$ with non-negative entries and $T_{ii}=0$
for all $i\in\{0,1,\ldots,n\}$ called the \emph{drive-time matrix}.
\item $n$ non-negative numbers $\{d_{i}\}_{i\in [n]}$ assigned to each node $z_{i}$ with $i\neq0$. These numbers are called \emph{initial
demands.}
\item A positive real number $C$, which is the capacity of the truck.
\end{enumerate}

The nodes of the graph $G$ represent a set of locations that a delivery truck may drive to. The node $z_0$ plays a special role, described below, and is called the \emph{depot}.
In general, $G$ need not be complete, \emph{i.e.} driving from $z_i$ to $z_j$ may be prohibited for certain pairs $(i,j)$. 

The matrix element $T_{ij}$ of the drive-time matrix $T$ is the time required for a truck to drive from $z_i$ to $z_j$ (this is why $T_{ii} = 0$). In a realistic scenario, the drive-time from $z_i$ to $z_j$ may be slightly different from the drive-time from $z_j$ to $z_i$, so $T$ is not required to be symmetric.

The initial demand $d_i$ is the volume of material  that starts at $z_i$. The units of demand need not be volume, but could instead be weight, monetary value, or any other positive real number associated with the material to be delivered. Volume is chosen here as the most appropriate for the problem at hand. All initial demand must be delivered to the depot, $z_0$, by a single truck with capacity $C$. When the truck stops at $z_i$ ($i\neq 0)$, it picks up as much demand as possible before departing to another node (either filling the truck to capacity or picking up all demand at $z_i$). This can result in part of the demand $d_i$ being left at $z_i$ when the truck first stops there, thus making it necessary for the truck to stop at $z_i$ again later to finish the delivery of $d_i$ (this is why the problem is said to allow ``split deliveries"). When the truck stops at the depot, it unloads all demand it is currently carrying before departing to another node. Demand that is not currently on the truck is called \emph{offboard demand}, and demand currently on the truck is called \emph{onboard demand}.

A truck route is a sequence of nodes $(z_{k_0}, z_{k_1}, \dots, z_{k_l})$ with $k_0 = k_l = 0$ (\emph{i.e.} routes must start and end at the depot). The total drive-time of a route can be written as
\begin{align}
    \text{time}(z_{k_0}, z_{k_1}, \dots, z_{k_l}) = \sum_{i=0}^{l-1}T_{k_ik_{i+1}}.
\end{align}
A demand-satisfying route is a route which results in 100$\%$ of the initial demand being delivered to the depot. The solution to the SDVRP is the demand-satisfying route with minimum total drive-time.

The SDVRP is a very commonly studied VRP in operations research, for which the usage of deep reinforcement learning is well-established in the literature. However, the problem we study in this work is significantly more complex than the SDVRP, as we will see in the next section. The model we use in this work is a generalized version of the model of \cite{kool2018attention}, built to address the generalized VRP described below.

\subsection{Generalized Vehicle Routing Problem}

Now we will define a more complex vehicle routing problem that generalizes many features of the SDVRP. We call this problem the generalized vehicle routing problem (GVRP). The GVRP more realistically models real world supply chain logistics, and in particular, the Aisin Corporation VRP is an instance of the GVRP. 

An instance of the GVRP is specified by the following data:
\begin{enumerate}
    \item A graph $G$ with $n$ nodes $z_{0},z_{1},\ldots,z_{n-1}$.
    \item An $n\times n$ matrix $T$ with non-negative entries and $T_{ii}=0$ for all $i\in\{0,1,\ldots,n-1\}$ called the \emph{drive-time matrix}.
    \item A finite set of indivisible boxes which serves as the demand for the problem. Each box has a volume and a required route.
    \item A positive real number $C$, which is the capacity of each truck.
    \item A positive real number $T_{\text{max}}$, which is the time-limit for daily supply chain operations.
\end{enumerate}

Just like in the SDVRP, $G$ represents the locations in the supply chain, and the entries of $T$ are times required for a truck to drive between pairs of nodes. But a crucial difference between the SDVRP and the GVRP is in the definition of demand. In the SDVRP, the initial demand is given as a set of non-negative real numbers $d_i$ representing the volume of demand that starts at $z_i$. In the GVRP, the demand is discrete; it is defined as a finite set of boxes, which cannot be further subdivided. Each box has two attributes:
\begin{enumerate}
    \item A volume, which is a positive real number.
    \item A required route, which is a finite sequence of nodes.
\end{enumerate}

Each box starts at the first node of its required route, and must travel to each node of its required route (in order) within the daily time limit. Such multi-leg delivery requirements can arise for several practical reasons in a real world supply chain environment. There may be capacity limitations at a box's destination node, causing that box to have to stop at a storage warehouse first before moving on to its destination. Also, the contents of some boxes may need to have an operation performed on them, such as an assembly procedure, before its final delivery. Another important reason for multi-leg delivery requirements is that a box may need to be sent back to its starting location empty, after the parts it was designed to carry have been delivered, so that the box can be used again the next day.

The daily time limit is another feature of the GVRP which is not present in the SDVRP. As a result, the optimization goal of the GVRP is different from that of the SDVRP. Instead of fixing the number of trucks and finding the demand-satisfying route with minimum total time, we impose a time limit and the goal is to find the demand-satisfying route that uses the minimum number of trucks and finishes within the time limit. 

The use of multiple trucks in the GVRP is another feature which distinguishes it from the SDVRP. This adds complexity to the problem in an obvious way, as optimal routes should now involve some form of cooperation between trucks. 

\section{Supply Chain Management Workflow}
\label{Algorithm}

The goal of this work is to apply reinforcement learning (RL) techniques to find commercially valuable solutions to GVRP instances. However, training an RL agent to make \emph{all} of the decisions that comprise a solution (both the truck routing decisions and the demand pickup and dropoff decisions) would be an extremely difficult task, as RL agents typically perform worse as the decision space grows larger \cite{dulac2015deep}. Due to the highly complex nature of the demand in the GVRP, the pickup and dropoff decisions generally come from a very large decision space. There can be thousands of boxes waiting at a node or on a truck, with many different individual routing requirements, so the number of options for which boxes to pickup or dropoff at a given stop can be astronomically large. The routing decisions on the other hand (which node a truck should drive to next), come from a much smaller decision space. If the graph $G$ contains $n$ nodes, then there are at most $n-1$ options for each routing decision. For this reason, our approach uses a trained RL agent to make the routing decisions, and sensible heuristic methods to make the pickup and dropoff decisions.

The basic workflow of our algorithm is
\begin{enumerate}
    \item Convert the discrete demand into a continuous \emph{tensor demand structure},
    \item Extract a ``good" subenvironment $S$ from the full supply chain environment,
    \item Use our trained RL agent to compute truck routes for the trucks operating in $S$,
    \item Use a heuristic method to compute pickups/dropoffs for the truck routes from step 3,
    \item Update the tensor demand of the full supply chain to account for the part of the demand that was delivered to its final destination in the solution for $S$,
    \item Repeat steps 2-5 until the demand of the full supply chain reaches 0.
\end{enumerate}

We refer to steps 2-5 as an \emph{iteration}. Each iteration produces a solution for one subenvironment. This solution consists of a set of truck routes, and instructions indicating which boxes each truck will pickup and dropoff at each stop along its route. To construct the full solution from these subenvironment solutions, we simply run all of the subenvironment solutions simultaneously. This iterative structure allows us to easily scale our method to solve problems with larger initial demand. Problems with more demand will simply require more iterations (\emph{i.e.} more subennvironments).

Each iteration consists of two phases: route-finding in Phase 1, pickup/dropoff-finding in Phase 2 using the routes from Phase 1 as input. There are two reasons for this two-phase approach, which will become clearer by the end of this section. The first reason is the split in how different decisions are made, described in the first paragraph of this section. The second reason is that we make the routing decisions using a simplified model of the demand. Good routing decisions cannot be made without at least some information about the demand, but using the true demand with all of its individual boxes complicates things significantly. We simplify the demand by combining all boxes that have the same required route into a ``box soup" which is infinitely divisible. This continuous approximation of the true discrete demand saves us the trouble of keeping track of thousands of individual boxes while making the truck routing decisions. But Phase 2 must then return to the true demand of the problem, and make pickup/dropoff decisions for each individual box at each stop for each truck.

The rest of this section describes each step of our algorithm in detail, assuming we have a trained RL agent to make the truck routing decisions in Step 3. The next section will describe the exact nature of the RL agent and its training process.

\subsection{Tensor Demand Structure}

The first step of our algorithm is to convert the discrete demand into continuous ``box soup". This naturally leads us to organize the demand into a set of higher rank tensors, or a \emph{tensor demand structure}.

In the SDVRP, all demand must be delivered to the same node, $z_0$. This gives the problem a \emph{vector demand structure}, \emph{i.e.} the offboard demand at time $t$ is given by a vector

\begin{align}
    D^t = (d_1^t, d_2^t, \dots, d_n^t).
\end{align}

In the GVRP, the offboard demand at time $t$ naturally takes the form of one or more higher rank tensors. This is easiest to see in the case of rank-2 demand. Rank-2 demand is any demand that is required to move along a sequence of nodes of length 2. For example, a box currently at $z_3$ that must move to $z_5$ contributes to the $(3,5)$ component of a rank-2 offboard demand tensor. More generally, the $(i,j)$ component of the rank-2 offboard demand tensor at time $t$, $D_{ij}^t$ (a matrix in the rank-2 case), is a non-negative real number whose value is the total volume of demand that is currently at $z_i$ and must to move to $z_j$ to complete its required route. This idea naturally generalizes to higher ranks. A box currently at $z_2$, which must first move to $z_4$ and then to $z_6$ to complete its required route contributes to the $(2,4,6)$ component of a rank-3 offboard demand tensor, $D_{246}^t$. For any instance of the GVRP, the initial demand can be organized into a set of offboard demand tensors of different ranks, whose components are the total volumes of demand that must move along specific sequences of nodes.

Each truck's onboard demand in the GVRP also takes the form of one or more higher rank tensors. Suppose truck $T$ arrives at $z_i$ and picks up one box of volume $V$, whose required route is $(i,j,k)$. Before being loaded onto the truck, this box contributes volume $V$ to the $(i,j,k)$ component of the rank-3 offboard demand tensor, $D_{ijk}$. After being loaded onto the truck, the box now contributes volume $V$ to the $(j,k)$ component of $T$'s rank-2 onboard demand tensor, $E^T_{jk}$. This allows us to keep track of the specific routing requirements of all demand on each truck. Note that all onboard demand tensors are 0 at the start of the day, before supply chain operations have commenced.

There is one more level of complexity in the demand structure. Often in real-world supply chains where the demand consists of small, intricately shaped, fragile parts, specially designed boxes are used to transport these parts from node to node. In the very common scenario where the supply chain has repeated operations, such as daily or weekly shipments of identical parts, these boxes get reused for each shipment. This means that those boxes need to return to their original locations after completing their required routes, thus forming \emph{cyclic} box routes. The simplest way to enforce this constraint is to add one more node (the start node) to the required route of any box that must return to its original location. This has the effect of increasing the ranks of some demand tensors by 1. But a more memory-efficient way is to distinguish between \emph{cyclic demand} and \emph{direct demand} (boxes which must be returned to their original locations, and those which do not), by defining a cyclic offboard demand $D^{\text{cyclic}}$ and a direct offboard demand $D^{\text{direct}}$. To account for the different routing requirements, there is a slight difference in how these tensors evolve as the demand gets delivered throughout the day, which will be described when we get to Step 3 of Phase 1.

\subsection{Subenvironment Search}
\label{sub_search}

The second step of our algorithm is to extract a subenvironment from the full supply chain environment. A subenvironment is a subset $S$ of the nodes of $G$, together with a subset $D_S$ of the total demand, and a set of $N$ trucks which are assigned to operate only within $S$ (stopping only at nodes in $S$). $D_S$ is fully determined by $S$: it consists of all demand whose required route lies fully inside $S$. The daily time limit for $S$ is equal to the daily time limit $T_{\text{max}}$ for the full supply chain. The reason we operate on subenvironments, instead of on the whole supply chain at once, is to make the routing decisions spaces smaller. When there are fewer nodes to drive to, the routing decisions become easier.

Here, we describe how ``good" subenvironments are chosen in Step 2. The goal is to choose a subenvironment that results in a large volume of demand being delivered. We estimate how much demand will be delivered in a subenvironment $S$ by $N$ trucks within time $T_{\text{max}}$ by using our trained RL agent to produce a solution for $S$, and looking at how much total volume is delivered by this solution. To find a solution for a subenvironment $S$, we run an \emph{episode} on $S$ (episodes are described below in Step 3). To find a good subenvironment, we test many subenvironments, running multiple episodes on each subenvironment and then choosing the subenvironment with the highest mean volume of delivered demand.

How do we decide which subenvironments to test? First we choose how many nodes we want in our subenvironment, $n^\prime$. Now we consider the initial offboard demand $D^0$. Assume for this example that $D^0$ consists only of rank-3 demand (this method readily generalizes to other ranks). Consider only the nonzero components of $D^0$. These are indexed by tuples of nodes
\begin{align*}
    (i_1, &j_1, k_1)\\
    (i_2, &j_2, k_2)\\
    &\vdots\\
    (i_u, &j_u, k_u)\\
\end{align*}
where $u$ is the number of distinct box routes in the initial demand. We start by randomly selecting one node tuple from the list, and then removing it from the list. 

Suppose that we select the tuple $(i_5, j_5, k_5)$. We then add these three nodes to our (initially empty) node subset, to form $A = \{i_5, j_5, k_5\}$. If $3 < n^\prime$,
we continue to randomly select node tuples. Suppose that we select $(i_4, j_4, k_4)$ next. We now add these nodes to $A$, to form the set $A=\{i_5, j_5, k_5, i_4, j_4, k_4\}$. We do not allow repeated elements (\emph{i.e.} $A$ is a set and not a multiset), so $A$ now contains \emph{at most} six elements. We continue this process of randomly selecting tuples of nodes and adding them to $A$ via set union, until we either run out of tuples or $|A| \geq n^\prime$. At this point, if $|A| = n^\prime$, we use $A$ as our test subenvironment. If $|A| > n^\prime$, we remove the subset added last. At this point, we have $|A| < n^\prime$, so we randomly add $n^\prime - |A|$ nodes.

This process gives us a node subset $A$ with $|A| = n^\prime$. We repeat this
process $k_{\text{num subsets}}$ times to produce $k_\text{num subsets}$ different subenvironments, and we run $k_\text{subset attempts}$ episodes on each subenvironment\footnote{This is small compared to the number of episodes we run on the chosen subenvironment in Step 3}. We then compute the mean volume of delivered demand for each subenvironment's $k_\text{subset attempts}$ episodes, and select the subenvironment with the highest mean volume of delivered demand.

\subsection{Phase 1: Route-finding}
\label{route-finding}

In the next step, we obtain a solution for the chosen subenvironment of step 2 by running a large number of episodes on this subenvironment, each producing a solution. We then choose the solution which leads to the largest volume of delivered demand. 

First we describe what an episode is. An episode is one full day ($T_{\text{max}}$) of supply chain operations, with the continuous tensor demand structure described above. An episode consists of a series of time-steps, with each time-step corresponding to a truck stopping at a node. Since there are multiple trucks operating in a single episode, each time-step has an associated \emph{active truck} and \emph{active node}. The active truck is the one making the stop, and the active node is the location of the active truck.

At each time step, three things happen:

\begin{enumerate}
    \item The active truck drops off any demand whose next required stop is the active node
    \item We invoke our trained RL agent to determine which node the active truck will drive to next
    \item The active truck picks up offboard demand from the active node according to a heuristic, and departs for its next node (chosen in 2)
\end{enumerate}

This continues until there is no more demand to deliver, or the daily time limit is reached. The primary complexity of this process is the bookkeeping associated with steps 1 and 3. Keeping track of where all of the demand is at all times is crucial for the accuracy of our computation.

To clarify, we provide an example in Table \ref{demand flow} of the demand flow for material with rank-3 required route, and a cyclic box-return constraint. The first column says which demand tensor component the material's volume contributes to after the event in the second column. 

\begin{table*}
\centering
\begin{tabular}{|c|c|}
\hline 
Component & Most recent event\tabularnewline
\hline 
\hline 
$D_{2\,4\,6}^{\text{cyclic}}$ & Material initially at node 2; must go to node 4, then node 6, then node 2 
%\vspace{3pt}
\\[1pt]
\hline 
$E_{4\,6\,2}^{m=1}$ & Picked up from node 2 by truck 1, next stop node 4\\[1pt]
\hline 
$D_{4\,6\,2}^\text{direct}$ & Dropped off at node 4 by truck 1; must go to node 6, then node 2\\[1pt]
\hline 
$E_{6\,2}^{m=2}$ & Picked up from node 4 by truck 2, next stop node 6\\[1pt]
\hline 
$D_{6\,2}^\text{direct}$ & Dropped off at node 6 by truck 2; must go to node 2\\[1pt]
\hline 
$E_{2}^{m=3}$ & Picked up from node 6 by truck 3, next stop node 2\\[1pt]
\hline 
0 & Dropped off at node 2 by truck 3, requirements fulfilled\tabularnewline
\hline 
\end{tabular}
\vspace{5pt}
\caption{An example of the flow of rank-3 cyclic demand from starting location to final location}
\label{demand flow}
\end{table*}

\subsection{Phase 2: Pickup and Dropoff Decisions}

Phase 2 is very simple compared to Phase 1. The inputs are the routes from Phase 1 for each truck in each subenvironment, the initial (discrete) demand, the daily time limit $T_{\text{max}}$, and the truck capacity $C$.

In Phase 2, we make pickup/dropoff decisions for each stop of each truck using the simple heuristics explained below. Crucially, these decisions are made at the individual box level, as opposed to at the ``box soup" level of Phase 1.

Each time a truck stops at a node, two things happen:
\begin{enumerate}
    \item Certain boxes on the truck are dropped off at the node.
    \item Certain boxes at the node are picked up by the truck.
\end{enumerate}

The heuristic used to determine which boxes to drop off is just a discrete version of the same heuristic used in Phase 1: we drop off any box whose next required stop is the current node. 

The heuristic used to determine which boxes to pick up is a bit more complicated, but still quite simple. Suppose the truck is at its $p$th stop, $z_{k_p}$, and the truck's remaining route is $(z_{k_{p+1}}, z_{k_{p+2}}, \dots, z_{k_l}$). We first pick up as many boxes as possible whose next destination is $z_{k_{p+1}}$. This means either fill the truck to capacity with these boxes, or continue loading these boxes until none remain. Then, if there is space on the truck to do so, load as many boxes as possible whose next destination is $z_{k_{p+2}}$. Continue in this fashion until either the truck is full, or there are no boxes remaining at $z_{k_p}$ whose next required stop is on the truck's remaining route.

There is one important subtlety to keep in mind here. We cannot allow any partial deliveries in Phase 2. In other words, we cannot allow any box to be moved from its starting location without fully completing its required route. This is because the subenvironment solutions generated by each iteration will all run simultaneously in the full solution. Therefore, a partial delivery in one iteration cannot be completed in another iteration without substantially complicating our algorithm. 

To eliminate partial deliveries in Phase 2, we simply add a subroutine at the end of Phase 2 which resets any partially delivered boxes to their starting locations, and removes them from the pickup/dropoff schedules, as if they were never moved.

\subsubsection{Total Demand Update}

Finally, in preparation for the next iteration, we must update the demand tensors that we constructed in Step 1, by subtracting any demand that was delivered to its final destination in the previous iteration. 

\section{Policy Neural Network for Routing Decisions}
\label{Neural Net}

The model we use for our neural networks is an encoder and decoder model directly inspired by that
of \cite{kool2018attention}. Their model is quite general: after training with REINFORCE \cite{reinforce-williams}, it performs well for numerous routing problems including the traveling salesman problem, basic vehicle routing problems, and the orienteering problem. However, it does not account for multiple trucks.
Moreover, there is no obvious way to incorporate a tensor demand structure into their model without a substantially new approach.

\subsection{Reinforcement Learning for Vehicle Routing Problems}

Before describing our neural networks, we first explain
how VRPs fit into the
framework of reinforcement learning, and also how an encoder
and decoder furnish a policy.

RL is often used in the context of a Markov decision process (MDP). In this setting, there is an environment that is in some state, and an agent who can perform certain actions which change the state of the environment. Given the environment state, the agent must choose an action from a set of allowed actions. This action results in a new environment state, and a reward for the agent. This process is repeated until some halting condition is met. The agent's goal is not to maximize the reward for a single action, but to maximize the long term reward, or \emph{return}.

The RL agent achieves its goal by learning a \emph{policy}, which is a conditional probability distribution over possible actions to take given an environment state. Given a state $s$ and an action $a$ from a set $A$ of potential actions, the agent learns to compute a quantity $\pi(a|s)\in [0,1]$ such that $\sum_{a\in A} \pi(a|s) = 1$. Given a state $s$, the agent can then sample from the policy $\pi$ to choose its next action. The agent's goal is to learn the optimal policy, \emph{i.e.} the one which maximizes the expectation value of the return.

In the context of the GVRP, The environment state $s$ consists of demand data (locations and required routes of all demand) and truck data (location and remaining capacity of each truck, index of the active truck). The action set $A$ is just the set of nodes that the active truck can drive to next.

\subsection{Encoder}\label{sec:encoder}
The encoder is used once at the beginning of each episode. Its purpose is to encode the data defining the problem into a set of high dimensional vectors, one for each node. The encoder we use is almost identical to the encoder of \cite{kool2018attention}, but with an adjustment to account for the higher rank demand.

The input data for the encoder contains information about the locations of the nodes, as well as information about the initial demand structure. The node locations can be fully captured by simple 2-dimensional coordinate vectors. The demand structure, however, is too complex to encode in its entirety, so we construct an abbreviated version as follows.

We define outgoing and incoming offboard demands at each node as
\begin{align}
    \delta^\mathrm{out}_i &= \sum_jD_{ij} + \sum_{jk}D_{ijk} + \sum_{jkl}D_{ijkl} + \dots,\label{eq:ingoing-outgoing-initial-demand}\\
    \delta^\mathrm{in}_i &= \sum_jD_{ji} + \sum_{jk}D_{jik} + \sum_{jkl}D_{jikl} + \dots\label{incoming demand},
\end{align}
where $D$ is defined to be the sum of both the cyclic and direct demands. Now we can define the input vector for node $z_i$ as the 4-dimensional vector
\begin{align}
    \overline{\mathbf{x}}_{i}
    =\mathbf{x}_{i} \oplus\left(\delta_{i}^{\text{in}},\delta_{i}^{\text{out}}\right),
\end{align}
where $\mathbf{x}_i$ is the 2-dimensional coordinate vector giving the location of node $z_i$, and $\oplus$ denotes concatenation. The set of vectors $\{\overline{\mathbf{x}}_1, \dots, \overline{\mathbf{x}}_n\}$ forms the input for the encoder. The first encoding layer is a linear map with bias from $\mathbb{R}^4$ to an encoding space $\mathbb{R}^d$ (we typically use $d = 64$),
\begin{align}
    \mathbf{h}^0_i = W^\mathrm{init} \overline{\mathbf{x}}_i + \mathbf{b}^\mathrm{init},
\end{align}
where $W^\mathrm{init}$ is a $d\times 4$ matrix and $\mathbf{b}^\mathrm{init}$ is a $d$-dimensional vector. Note that the same encoding map is applied to every every vector in the input set, \emph{i.e.} $W^\mathrm{init}$ and $\mathbf{b}^\mathrm{init}$ do not depend on $i$.

In addition to the initial layer, our encoder consists of two more layers: an attention layer (\ref{eq:enc-layer-att})
followed by a feedforward layer (\ref{eq:enc-layer-ff}), given by

\begin{align}
    \tilde{\mathbf{h}}^{l-1}_{i} &= 
    \mathrm{BN}\left(
    \mathbf{h}^{l-1}_{i}  + \mathrm{MHA}\left(\mathbf{h}^{l-1}_{i}\right) 
    \right)\label{eq:enc-layer-att},
    \\\mathbf{h}^{l}_{i} &=
    \mathrm{BN}\left(\tilde{\mathbf{h}}^{l-1}_{i}  + \mathrm{FF}\left(\tilde{\mathbf{h}}^{l-1}_{i}\right) \right).\label{eq:enc-layer-ff}
\end{align}

Here, $\mathrm{BN}$ is batch normalization \cite{ioffe2015batch}, $\mathrm{FF}$ is a feed-forward layer, and MHA is a multi-head attention layer, which we describe in detail below. The feed-forward layer has a hidden dimension $d_\mathrm{ff}$ and is made up of a linear layer with bias which maps $\mathbb{R}^d \to \mathbb{R}^{d_\mathrm{ff}}$, followed by a ReLU, a dropout layer, and another linear layer with bias to $\mathbb{R}^d$.

\subsubsection{Multi-head Attention Mechanism}

The function MHA appearing in (\ref{eq:enc-layer-att}) is a multi-head attention mechanism which is identical to that of \cite{kool2018attention}. Variations of the MHA layer are used in both the encoder and the decoder, with a modification described below for dealing with a tensor demand structure.

We start with the set of output vectors $\mathbf{h}_{1}, \ldots, \mathbf{h}_{n}\in\mathbb{R}^{d}$ from the previous layer. For each of these vectors, we compute vectors called
queries, keys, and values. In a single-head attention mechanism, one of each type of vector (query, key, value) is computed for each input vector $\mathbf{h}_{i}$,
\begin{align}
    \mathbf{q}_{i} &= M^\mathrm{query}\,\mathbf{h}_{i} \in\mathbb{R}^{\alpha},\\
    \mathbf{k}_{i}&=M^\mathrm{key}\,\mathbf{h}_{i}\in\mathbb{R}^{\alpha},\\
    \mathbf{v}_{i}&=M^\mathrm{value}\,\mathbf{h}_{i}\in\mathbb{R}^{d}.
\end{align}
$M^\mathrm{query}, M^\mathrm{key}, M^\mathrm{value}$ are linear layers, and $\alpha$ is any positive integer.

For a multi-head attention mechanism, we have a positive integer $n_\mathrm{heads}$ which, in our case, we typically take to be eight. Each attention head, labeled by $s\in [n_\mathrm{heads}]$, now gets its own query, key, and value map,
\begin{align}
    \label{eq:q-heads-no-source}
    \mathbf{q}_{s\,i} &= M_s^\mathrm{query}\,\mathbf{h}_{i} \in\mathbb{R}^{\alpha},\\
    \label{eq:k-heads-no-source}
    \mathbf{k}_{s\,i}&=M_s^\mathrm{key}\,\mathbf{h}_{i}\in\mathbb{R}^{\alpha},\\
    \label{eq:v-heads-no-source}
    \mathbf{v}_{s\,i}&=M_s^\mathrm{value}\,\mathbf{h}_{i}\in\mathbb{R}^{\beta},
\end{align}
where $\beta$ can be any positive integer. The linear maps $M$ are learned during training.

Next, a \emph{compatibility} is computed for every query-key pair (for each head), via the standard dot product,
\begin{equation}
\label{eq:basic-compat}
    u_{s\,ia}=\frac{1}{\sqrt{\alpha}} \mathbf{q}_{s\,i}\cdot \mathbf{k}_{s\,a}.
\end{equation}
Then, for each head $s$ and node $i$, we pass the compatibilities $\{u_{s\,ia}\}_{a\in [n]}$ through a softmax to obtain
\begin{equation}
    \label{eq:softmax}
    \rho_{s\,ia}=\frac{\exp(u_{s\,ia})}{\sum_{b}\exp(u_{s\, ib})}\in\mathbb{R}.
\end{equation}
The softmax compatibilities are then used as coefficients in a linear combination of value vectors,
\begin{equation}
    \label{eq:new-nodes-att}
    \mathbf{g}_{s\, i}=\sum_{a}\rho_{s\, ia}\mathbf{v}_{s\, a}.
\end{equation}

The next step is to merge the data from the $n_\mathrm{heads}$ attention heads, by simply concatenating the outputs from each
head,
\begin{align}\label{head merge}
     \mathbf{g}_i = \mathbf{g}_{1\, i} \oplus \ldots \oplus  \mathbf{g}_{n_\mathrm{heads}\, i}.
\end{align}
Finally, we use a learned linear map with bias to map these vectors from dimension $n_\mathrm{heads} \beta$ to dimension $d$ to produce vectors $\mathbf{h}^\prime_{i} \in \mathbb{R}^d$, one for each node $i\in [n]$.

\subsection{Decoder}

After running the encoder once at the start of an episode, the decoder is then used once at every time step (after dropping off demand from the active truck) to decide which node the active truck will drive to next. Unlike the encoder, we make significant modifications to the decoder of \cite{kool2018attention}, in order to allow for multiple trucks.

The output from the encoder is a set of $n$ vectors $\{\mathbf{h}^l_i\}_{i\in [n]}$, one for each node. To compute a probability distribution over the other nodes (which we then sample to decide the next node), we ``decode'' these encoded nodes along with
some information about the current state of the environment.

The state of the environment, at the time we use the decoder to make a routing decision, consists of the following data:
\begin{itemize}
    \item the current on-board demands $E^m$,
    \item the current off-board demands $D^{\mathrm{direct}}$, $D^\mathrm{cyclic}$,
    \item the encoded nodes $\{\mathbf{h}_i\}_{i\in [n]}$, which were encoded once at the start of the episode,
    \item the remaining capacity of each truck,
    \item the active truck $m_{\star}$,
    \item the active node $z_\star$,
    \item the next expected nodes for each passive truck and the time until those trucks arrive.
\end{itemize}
Ideally, we would pass all of this data to the decoder. The difficulty is that there is too much demand data, so we must abbreviate the demand data to include only its most immediately relevant features, much like for the encoder.

\subsubsection*{Demand Information}

Since the demand tensor indices refer to nodes, a natural way to condense the demand data and pass it to the decoder is to reduce each demand tensor to rank-1 by summing over all but one index, and then concatenate its components with the corresponding encoded node.

More precisely, we use the outgoing offboard demand $\delta_i^\mathrm{out}$ defined in \eqref{eq:ingoing-outgoing-initial-demand}, and define a consolidated version of each truck's onboard demand as
\begin{equation}
    \label{eq:onboard-myopic}
    \epsilon^m_i = E_i^m + \sum_j E^m_{ij} + \sum_{jk} E^m_{ijk} + \dots 
\end{equation}
In words, $\epsilon^m_i$ is the total volume of demand on truck $m$ whose next required stop is node $i$.

Now, let $m_\star$ be the active truck index. We then modify the encoded nodes as follows
\begin{equation}
    \label{eq:encoded-extended-for-decoder}
    \overline{\mathbf{h}}_{i} = \mathbf{h}_{i} \oplus  \left(\delta_i^\mathrm{out}, \epsilon_{i}^{m_\star}, \epsilon_{i}^{1},
    \ldots,  \epsilon_{i}^{m_\star-1}, \epsilon_{i}^{m_\star+1}, \ldots, \epsilon_{i}^{N}  \right),
\end{equation}
where $\oplus$ is concatenation.

\subsubsection*{Additional Contextual Information}

In the previous subsection, we explained how we pass the encoded nodes and demand information to the decoder: we consolidate the demand data into a vector for each node, and then concatenate these vectors with the corresponding encoded node. But the remaining pieces of environment data -- the remaining capacity of each truck, the active truck index, the active node index, and the next node and arrival time for each truck -- are associated with trucks. Therefore, we need a different way to pass this information to the decoder.

To do this, we follow the model of \cite{kool2018attention} and introduce an extra node called the \emph{context node}. The context node will contain relevant information about the status of each truck. Let $m_\star$ be the active truck index and define
\begin{equation}
    \label{eq:cap-ctx}    \mathbf{C}=\left(C_{m_\star},C_{1},C_{2},\ldots,C_{m_\star-1},C_{m_\star+1},\ldots,C_{N}\right),
\end{equation}
where $C_k$ is the remaining capacity for truck $k$. $\mathbf{C}$ will form the first
part of our context node.

The second part of the context node will contain information about which node each truck will arrive at next. Let $k_{m_\star}$ be the active node index, and let $(k_1, \dots, k_{m_\star - 1}, k_{m_\star + 1}, \dots, k_N)$ be the next node indices for all passive trucks. Ideally, we would like to append to the context node
\[
\left(h_{k_{m_\star}},h_{k_{1}},h_{k_{2}},\ldots,h_{k_{m_\star-1}},h_{k_{m_\star+1}},\ldots,h_{k_{N}}\right),
\]
but this would be extremely expensive: the context vector would pick up $d N$ dimensions just from these components. To avoid this, we take the view that the passive trucks are less important than the active truck, and we use a feedforward neural network $f$, consisting
of a linear layer, ReLU nonlinearity, and then another linear layer, to reduce the dimension of the passive nodes (typically to four dimensions). We therefore define
\begin{multline}
    \label{eq:node-ctx}
    \mathbf{H} =\left(h_{z_{m_\star}},f\left(h_{z_{1}}\right),f\left(h_{z_{2}}\right),\ldots, \right) \\ \left(
    f\left(h_{z_{m_\star -1}}\right),f\left(h_{z_{m\star +1}}\right),\ldots,f\left(h_{z_{N}} \right) \right),
\end{multline}
and $\mathbf{H}$ will form the second part of our context node.

The third and final piece of the context node will contain information about how much time until each passive truck
arrives at its next scheduled destination. Let $t_m$ denote the time until truck $m$ arrives at its next stop. Note that we can have $t_m = 0$ if truck $m$ has the same arrival time as the active truck. Now we define
\begin{equation}
    \label{eq:time-ctx}
    \mathbf{T}=\left(t_{1},\ldots t_{m_\star-1},t_{m\star +1},\ldots,t_{N}\right).
\end{equation}
Note that the last $N-1$ components of $\mathbf{H}$ (equation (\ref{eq:node-ctx}))
correspond to the same trucks with the same order as the components of $\mathbf{T}$. This consistency is very important. With different events, the same components may correspond to different trucks, but for any given event, the components of $\mathbf{C},\mathbf{H},$ and $\mathbf{T}$ line up in the same way.

We finally define the \emph{context node}:
\begin{equation}
    \label{eq:ctx-node}
    \mathbf{h}_\mathrm{ctx} = \mathbf{H}\oplus\mathbf{T}\oplus\mathbf{C}.
\end{equation}

\subsubsection*{Decoder Structure}
The first layer of the decoder has exactly the same structure as the encoder layer of (\ref{eq:enc-layer-att}) and (\ref{eq:enc-layer-ff}), and takes as input the modified nodes of equation (\ref{eq:encoded-extended-for-decoder}). The only differences between this layer and the encoder layer are that the nodes have a greater dimension due to the modifications in equation (\ref{eq:encoded-extended-for-decoder}), and one additional modification explained
below. We denote the output as $\{\overline{\mathbf{h}}^1_{i}\}_{i\in [n]}$.

The additional modification is that, following \cite{kool2018attention},
we adjust the computation of keys and values in equations \eqref{eq:k-heads-no-source} and \eqref{eq:v-heads-no-source}
by adding \emph{source terms}:

\begin{align}
    \label{eq:k-heads-vec-source}
    k_{s\,i}&=M_s^\mathrm{key}\,\mathbf{h}_{i} + \mathbf{u}_s^\mathrm{key, out}\,\delta_i^\mathrm{out} + \mathbf{u}_s^\mathrm{key, in}\,\delta_i^\mathrm{in},\\
    \label{eq:v-heads-vec-source}
    v_{s\,i}&=M_s^\mathrm{value}\,\mathbf{h}_{i} + \mathbf{u}_s^\mathrm{val, out}\,\delta_i^\mathrm{out} + \mathbf{u}_s^\mathrm{val, in}\,\delta_i^\mathrm{in}.
\end{align}
Here, $\delta_i^\mathrm{out}$ and $\delta_i^\mathrm{in}$ are the outgoing and incoming demands at node $i$ defined in \eqref{eq:ingoing-outgoing-initial-demand} and \eqref{incoming demand}, respectively, evaluated at the time of decoding. $s$ indexes the attention heads, $i$ indexes the nodes, and the various $\mathbf{u}$'s are learned vectors with the same dimension as the object on the left hand side of the equations they appear in; for example, $\mathbf{u}_s^\mathrm{key,out}$ is a vector with the same dimension as the keys which is $\alpha$ as specified in equation \eqref{eq:k-heads-no-source}. This modification of keys and values helps to convey the current state of the demand
directly to the decoder.

After the initial decoder layer, we do a second attention layer with the same number of heads $n_\mathrm{heads}$, but acting on $n+1$ nodes: the $n$ nodes $\{\overline{\mathbf{h}}^1_{i}\}_{i\in [n]}$
from the first decoder layer, and the one context node \eqref{eq:ctx-node}. The purpose of this layer is only to produce a new transformed context vector. For this attention layer, following \cite{kool2018attention}, the only node we construct query vectors for is the context node, and we do not construct key and value vectors for the context node. In other
words, for each head we compute one query (for the context node), $n$ keys, and $n$ values (for all other nodes). In equations \eqref{eq:basic-compat}-\eqref{head merge}, the index $i$ only takes on a single value, the value which labels the context node. We use sources $\delta_i^\mathrm{out}, \delta_i^\mathrm{in}$, just as in equations \eqref{eq:k-heads-vec-source} and \eqref{eq:v-heads-vec-source}.
Moreover, for this layer we only compute MHA as described at the end of section \ref{sec:encoder}. This is a pure attention layer (as opposed to and encoder layer which uses equations \eqref{eq:enc-layer-att} and \eqref{eq:enc-layer-ff}).

The output of the second decoder layer is one new context vector $\mathbf{h}_\mathrm{ctx}^\prime$
with dimension $d_\mathrm{ctx}$. This new context vector is then used for a third and final layer which only uses one attention head. Once again, a query vector $q_\mathrm{ctx}$ is computed only for $\mathbf{h}_\mathrm{ctx}^\prime$, and keys are computed for the encoded nodes
$\overline{\mathbf{h}}^1_{i}$. There is no need to compute value vectors in the final layer, since we do not need to generate a new set of vectors to go into another layer (we want this layer to return a probability distribution over nodes). We will use the compatibilities directly to generate a distribution over nodes.

For this last layer, we compute compatibilities in the usual way except that we regulate with $\tanh$ and we allow for \emph{masking}:
\begin{equation}
    u_{i}=
    \begin{cases}
    A \tanh(q_{\text{ctx}}\cdot k_{i}) & \text{if node \ensuremath{i} is allowed}\\
    -\infty & \text{otherwise}
    \end{cases},
\end{equation}
where $A$ is a hyperparameter that we take to be 10. The idea is that we can block certain nodes for the active truck to drive to if we know, for some reason, that doing so is a poor choice. 

Finally, the $u_i$ are converted to probabilities with a softmax layer, and these probabilities are interpreted as the values of the policy: the probability
of selecting node $i$ for the active truck's next destination:
\begin{equation}
    \label{eq:policy-decoder-out}
    \pi(i) = \frac{e^{u_i}}{\sum_{j=1}^n e^{u_j}}.
\end{equation}

\subsection{Incorporating Tensor Demand Structure}

Thus far, all demand information passed to the attention mechanism has been consolidated into rank-1 objects. None of the higher rank demand data is accessible to our RL agent at this point. However, for any realistic problem instance, attempting to convey the entirety of the demand data in its exact form would consume far too much memory. Therefore, we have implemented a method for passing rank 2 demand data to the attention mechanism. This still falls short of giving the RL agent complete information about the state of the environment, but is an improvement over only using rank-1 data. The method we use is called \emph{dynamical masking}.

\subsubsection{Dynamical Masking}

A natural way to pass a rank-2 consolidated demand tensor $D_{ij}$ to the attention mechanism is to identify an object in the attention mechanism pipeline that makes use of two different node indices $i$ and $j$, and then multiply this object by some function of $D_{ij}$. The part of the attention mechanism that involves two nodes is the dot product evaluation between keys and queries. To incorporate a demand tensor $D_{ij}$, we can replace the dot product with
\begin{equation}
    \label{eq:dynamical-masking-simple}
   \frac{1}{\sqrt{\alpha}} G_{ij} \, \mathbf{q}_i \cdot \mathbf{k}_j ,
\end{equation}
where $G$ is some function of $D$. This approach has the advantage that it allows the network to learn to exaggerate or suppress certain compatibility values in certain demand environments.

Our $G$ contains four terms, each with a trainable coefficient. The first is a constant term which we set to 1. This term gives the basic dot product compatibility, as in (\ref{eq:basic-compat}). The next term is a matrix $M$ which acts as a mask $M_{ij} = 1$ if $D_{ij} > 0$ and $M_{ij} = -\infty$ if if $D_{ij} = 0$. These two terms are used in the method of \cite{kool2018attention}. The third term is $\log D_{ij}$. This term behaves like a mask in that $\log D_{ij}\to -\infty$ as $D_{ij}\to 0$, but also exaggerates compatibility when $D_{ij}$ is large. Finally, the fourth term is simply $D_{ij}$ itself, which makes the entire expression more sensitive to changes in large values of $D_{ij}$. 

Additionally, each attention head $s$ has its own trainable $G_{ij}^s$. So, altogether, dynamical masking is achieved by
\begin{equation}
    \label{dynamcal-masking}
    G^s_{ij} = A^s_\mathrm{basic} + A^s_\mathrm{mask} M_{ij} + A^s_\mathrm{log} \log D_{ij} + A^s_\mathrm{lin} D_{ij},
\end{equation}
with the four coefficients $A$ being the trainable parameters.

\section{Training Methodology}
\label{Training}

The method we use to train our RL agent includes two parts:
\begin{enumerate}
    \item an RL algorithm which adjusts the model weights to optimize the value of a cost function
    \item a method for generating synthetic training data
\end{enumerate}
In this section, we describe both parts of the training process.

\subsubsection*{REINFORCE Implementation}
\label{sec:reinforce}

As in \cite{kool2018attention}, we use a modified version of REINFORCE \cite{reinforce-williams,Sutton1998} to train our agent. REINFORCE is a policy-gradient RL algorithm, which means that instead of estimating the ``value'' of actions in various states and training to find actions with high estimated value, we work directly with a parameterized policy and vary the parameters $\theta$ to minimize the \emph{cost} from an episode. 

Our REINFORCE variant defines actions differently from most typical implementations. Since it is very difficult to evaluate how good or bad any single routing decision is, we treat an episode's entire series of routing decisions, $(\xi_1, \xi_2, \ldots \xi_k)$ where each $\xi_i$ is a node, as a single action. Since REINFORCE updates policy weights using the log probabilities of actions, and we define an action as the complete sequence of routing decisions in an episode, we simply use the sum of the log probabilities of each individual routing decision for the policy update step. 

\begin{figure}
    \centering
    \includegraphics[trim = 5 0 0 0, clip, scale = 0.56]{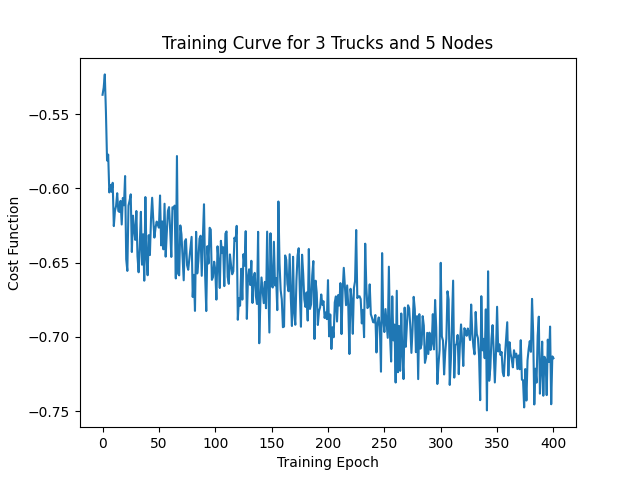}
    \caption{An example training curve showing the cost function, Eq. (\ref{eq:cost}), over 400 training epochs on subenvironments containing three trucks and five nodes.}
    \label{training_curve}
\end{figure}

Since the entire episode can be regarded as a single action, we only need to define a cost function $F(\xi)$ for a full route $\xi$. Typical VRPs can use total driving time or distance as a cost function to minimize. However, our routing problem has a time constraint $T_\mathrm{max}$ and it is not guaranteed that all demand will be fulfilled. To address this, we define \emph{demand coverage} as the percentage $\eta(\xi)$ of initial demand volume that is eventually fulfilled by route $\xi$. We then define the cost function as simply
\begin{equation}
    \label{eq:cost}
    F(\xi) = -\eta(\xi),
\end{equation}
which is then minimized in training. An example training curve is shown in Fig. \ref{training_curve}.

We have also implemented a \emph{baseline} policy for faster convergence. The baseline policy uses the same model as the primary policy, and starts with exactly the same parameters, $\theta_{\text{BL}} = \theta$. After each epoch, the subroutine \texttt{baseline\_test()} decides whether or not to update the baseline policy parameters $\theta_\text{BL}$ according to the following rule. If the fraction of episodes where the baseline policy achieves lower cost than the primary policy is greater than 50\% in all of the last 10 epochs, or this fraction is greater than 70\% in the last epoch, then we update $\theta_\text{BL}$. On epochs where \texttt{baseline\_test()} returns \texttt{true}, $\theta_\text{BL}$ is updated to be equal to the primary policy parameters $\theta$.

\begin{algorithm}
\caption{REINFORCE variant for GVRP}\label{alg:reinforce-variant}
\begin{algorithmic}
\STATE Input: Parameterized policy $\pi$
\STATE Input: Integers \texttt{num\_epochs, batch\_size, batches\_per\_epoch}
\STATE Input: Initial parameter $\theta$
\STATE $\theta_{\text{BL}} \gets \theta$
\FOR {$e =1, \ldots,$ \texttt{num\_epochs}}
    \FOR{$b =1, \ldots,$ \texttt{batches\_per\_epoch}}
        \STATE $\xi \gets$  (\texttt{batch\_size} many episodes from $\pi(\theta)$)
        \STATE $\xi_\text{BL} \gets$  (\texttt{batch\_size} many episodes from $\pi(\theta_\text{BL})$)
        \STATE $\nabla J \gets \texttt{batch\_mean}\newline~~~\quad\quad\left[
            \big(F(\xi) - F(\xi_\text{BL})\big) \nabla_\theta \left(\sum_{i=1}^k \log\pi(\xi^i, \theta) \right)
        \right]$
        \STATE $\theta \gets \textrm{descent}(\theta, \nabla J(\theta))$
    \ENDFOR
    \IF {\texttt{baseline\_test()}}
        \STATE $\theta_\text{BL} \gets \theta$
    \ENDIF
\ENDFOR
\end{algorithmic}
\end{algorithm}
Our modified version of REINFORCE is shown in Algorithm \ref{alg:reinforce-variant}. To reiterate the definition of actions discussed above, notice the summation $\sum_{i=1}^k \log\pi(\xi^i, \theta)$ appearing in algorithm \ref{alg:reinforce-variant}. This is the sum of the log probabilities of each routing decision in the episode, computed by the neural network described in the previous section. As explained above, we use this form because it is equal to the log of the product of the probabilities, and the product of probabilities gives the probability of the whole route (or the probability of the episode's single action). $k$ is the number of stops in the episode's route. The expression involving the gradient of sum of the log probabilities is evaluated for each batch entry, and these values are then averaged over the entire batch. The intuition is that this should increase the probability of routes with lower cost and decrease the probability of routes with higher cost.

\subsubsection*{Environment Generation}
\label{sec:enc}
In order to train a model that can solve a wide range of problem instances, we randomly generate batches of synthetic supply chain environments and use these as training data. A supply chain environment consists of a graph representing the set of locations, and a set of initial demand tensors. We must randomly generate both of these for each synthetic training environment.

We randomly generate graphs using a single parameter, a timescale $\tau$, that represents the scale of drive times we want in our environment. To generate an $n$-node graph, we simply choose $n$ random points in the unit square, and then multiply the coordinates of all $n$ points by $\tau$. The drive times are then computed as the pairwise distances between points.

Randomly generating initial demand tensors is a bit more involved. An arbitrary tensor of the correct rank and dimension will, in general, not be acceptable. For example, a component like $D^\mathrm{cyclic}_{232}$ must be excluded, since this would imply a required box route $(z_2, z_3, z_2, z_2)$. For a rank $r$ demand tensor, we start by randomly choosing a subset of nodes which are allowed to be the first node of a box route, a subset which are allowed to be the second, and so on up to a subset of nodes which are allowed to be the final ($r$th) node of a box route. We then generate a mask tensor whose components are 1 only for those box routes which are allowed according to these node subsets, and contain no repeated nodes. We then further mask components randomly with some given probability. This is meant to create instances which are more representative of real problem instances, where the demand structure does not have all-to-all connectivity. We then randomly assign the unmasked components a value in $(0,1]$, and multiply the entire tensor by a positive number which represents the scale of demand we want in our synthetic environment. We repeat this process once for each demand rank $r$ we want our synthetic environment to contain. Our synthetic training environments contain demand up to rank 3.

\section{Results}
\label{Results}

We apply the algorithm described above to solve a special case of the GVRP, which we call the Aisin VRP. The Aisin VRP arises in the daily supply chain operations of Aisin Corporation in Japan's Aichi Prefecture. In this section we describe the Aisin VRP, and then evaluate the performance of our algorithm in solving this problem.

\subsection{Aisin VRP}

In the Aisin VRP, the graph $G$ consists of 21 nodes, representing 21 different supply chain locations. The drive-time matrix $T$ has entries that range from two minutes up to about two hours. There are approximately 340,000 individual boxes to deliver, with a total volume of approximately 11,500m$^3$. Each box has one of 107 unique required routes. The trucks have volume capacity $C = 30$m$^3$, and the daily time limit is 16 hours. To solve this problem, Aisin Corporation relies on a team of logistics experts using intuition and experience to plan the truck routes and pickups/dropoffs.

Our goal is to produce solutions to this problem which are on par with or better than Aisin Corporation's current best solution. Aisin's previous best solution uses 142 trucks to complete this supply chain task in 16 hours. We use this number as our performance benchmark. 

There is an important caveat to mention here. There are several constraints that appear in the real-world version of this problem, which we have not imposed in our solution. First, our algorithm does not include a minimum time for which a box must stay at an intermediate node along its required route. This means that boxes in our solution can be picked up for the next leg of their journey immediately after they are dropped off, which is somewhat unrealistic. Second, the trucks have a weight capacity in addition to a volume capacity, but our algorithm does not impose a weight constraint. Finally, realistic truck routes must start and end at the same node so that the truck driver can park their car at the start node at the beginning of the shift, and get back in their car to drive home at the end of the shift. We have not imposed a cyclic truck route constraint. Even without these constraints, the solutions we find still represent a big step towards a commercially viable RL approach for realistic VRPs.

\subsection{Performance}

The results presented in this section were obtained using the algorithm described above, with teams of three trucks operating in 5-node subenvironments. The initial encoding dimension is 64, and both the encoder and decoder MHA layers have eight attention heads. We trained our model for 400 epochs using Adam optimization with an exponentially decaying learning rate. The learning rate started at 0.05 and decreased by a factor of 0.9 after each epoch until reaching its minimum value of 2$^{-14}$. The subenvironment search routine described in Sec. \ref{sub_search} tested 20 different subenvironments in each iteration, and ran 20 test episodes on each test subenvironment. After choosing a subenvironment, we ran a batch of 500 episodes on the chosen subenvironment.

To get a sense for how our algorithm performs as a function of the total amount of demand in the supply chain, we consider nine problem instances each with a different demand scale. Each problem instance was extracted from the full-scale Aisin VRP by considering only a subset of the total demand of that problem. As fractions of the total demand of the Aisin VRP, these problem instances contain 2.5$\%$, 5$\%$, 10$\%$, 15$\%$, 20$\%$, 25$\%$, 50$\%$, 75$\%$, and 100$\%$ of the total demand volume. 
\begin{figure}%[H]
    \centering
    \includegraphics[trim = 16 0 0 0, clip, scale = 0.58]{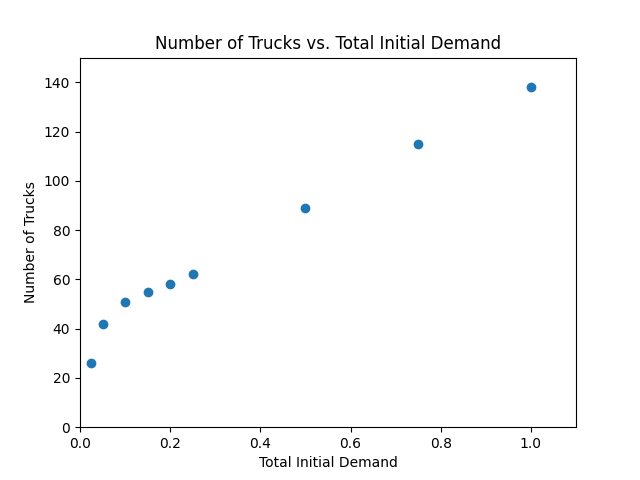}
    \caption{Number of trucks required as a function of total initial demand. Total initial demand is given as a fraction of the total demand of the Aisin VRP. Note for the full-scale problem, our algorithm finds a solution using 138 trucks, thus outperforming Aisin's 142-truck solution. However, as seen on the left side of the plot, the algorithm is less effective with a small amount of demand spread over many nodes.}
    \label{Trucks_vs_Demand_plot}
\end{figure}
What is clearly visible in Fig. \ref{Trucks_vs_Demand_plot} is that this algorithm delivers significantly less demand volume per truck when faced with a smaller amount of demand spread across the same number of nodes. This is not surprising, since a small amount of demand spread across many nodes means that trucks will be less full on average, and therefore must do more driving to deliver the same amount of demand. 
\begin{figure}%[H]
    \centering
    \includegraphics[trim = 800 0 800 0, clip, scale = 0.1]{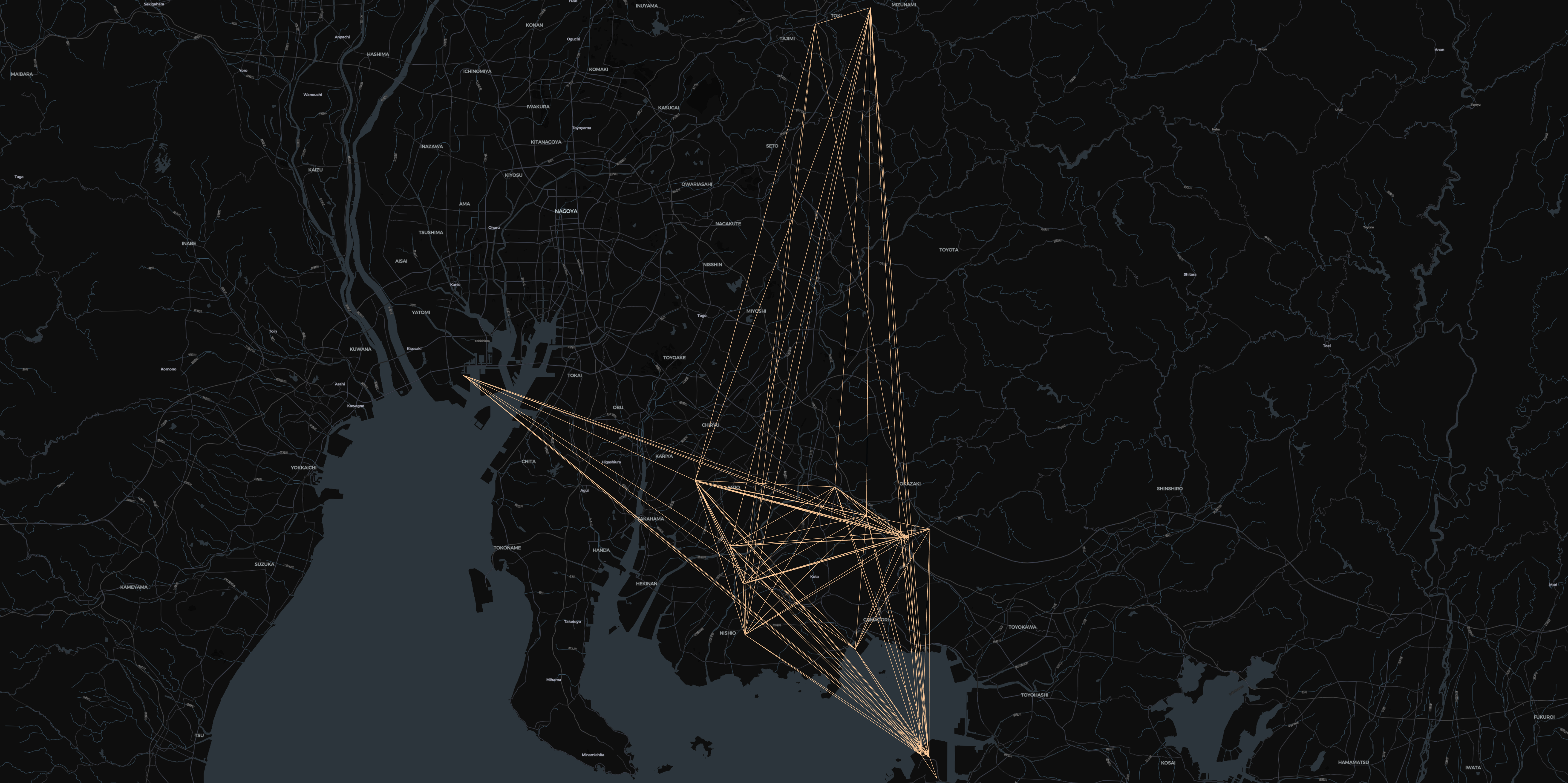}
    \caption{Truck-routing connectivity graph for the 138-truck solution obtained using the algorithm described in this paper. The underlying map shows an area of approximately 85km by 85km of Nagoya, Japan.  Note that this figure only shows which edges are used in the solution, and does not show the demand flow along each edge, which direction trucks are driving, or timing details.}
    \label{Aichi_image}
\end{figure}
Let us focus now on the full-scale Aisin VRP, represented in Fig. \ref{Trucks_vs_Demand_plot} by the rightmost data point. Our algorithm was able to solve this problem using only 138 trucks, thus outperforming the 142-truck solution currently used by Aisin. The graph edges used in this solution are shown in Fig. \ref{Aichi_image}.

\begin{figure}%[H]
    \centering
    \includegraphics[trim = 16 0 0 0, clip, scale = 0.58]{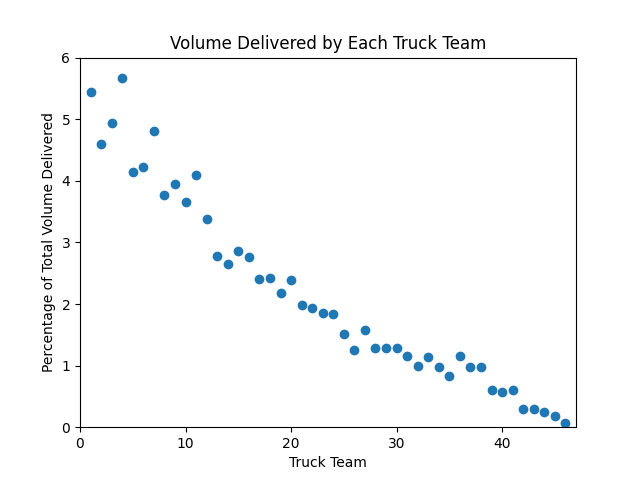}
    \caption{The percentage of the total initial demand volume delivered by each 3-truck team. The first teams deployed are the most efficient due to the abundance of demand available to them.}
    \label{Volumes_each_team}
\end{figure}

Fig. \ref{Volumes_each_team} shows the volume of demand delivered by each truck team in the 138-truck solution. Here we can see the same effect that is visible in Fig. \ref{Trucks_vs_Demand_plot}. The truck teams that are deployed in later iterations deliver less volume per truck. This is because the later truck teams operate in an environment with small amounts of demand spread over all 21 nodes, so it is much harder for them to be efficient.

\section{Conclusion}
In this work, we have developed an algorithm that uses deep reinforcement learning with an attention-based model to solve realistic VRPs with multiple trucks and multi-leg routing requirements. These problems are made even more complex by the presence of a box-return constraint making all box routes cyclic. An important feature of our algorithm is that it can produce solutions to problems of arbitrary size without having to scale up the size of the model. It does this by using an iterative procedure in which each iteration produces a solution to one small sub-problem, so larger problems can be solved by simply doing more iterations of the same algorithm using the same model. We tested our algorithm on a real supply chain environment of Aisin Corporation, and found that our algorithm outperformed Aisin's solution for a slightly simplified version of this problem.

As mentioned in the previous section, it is important to point out that the solutions we obtain using the algorithm described in this paper cannot be immediately deployed. Our algorithm, despite incorporating many realistic aspects of Aisin supply chain environments, still does not include every constraint that Aisin must contend with. It is likely that our algorithm will need to further improve in order to outperform Aisin's solutions after adding all necessary constraints. Future work will involve incorporating more real-world constraints into our approach, and adjusting our algorithm to handle the added difficulty.

While the results presented here are promising, there is still quite a bit of room for improvement. The most noticeable weakness of our algorithm is that we deploy truck teams into the supply chain environment in series, updating the total demand each time by subtracting any demand that was delivered by the previous truck team. The result of this, which is apparent from Figs. \ref{Trucks_vs_Demand_plot} and \ref{Volumes_each_team}, is that the trucks deployed later in the workflow are very inefficient, since they only have small bits of demand available to pick up but still must drive the same distances to make deliveries. This problem could be mitigated by using an approach that deploys all trucks simultaneously. This way, all trucks would have access to the full scale of the problem demand at the moment they are deployed. Such an approach should be a goal of future work on this problem.

Another part of our algorithm that can be improved is the pickup heuristic used in Step 3 of the route-finding routine described in Sec. \ref{route-finding}. During training, we run episodes that use the updated model to make the routing decisions, and this heuristic to make the pickup decisions. This means that the learning process of our model is highly dependent on the details of the pickup heuristic. A suboptimal pickup heuristic will lead to suboptimal route-finding. An advanced approach would use a trained neural net to make all decisions, both routing and pickup, since ML tends to work best when hand engineering is kept to a minimum. However, this may be very difficult due to the very large number of boxes in the problems we consider. Another option is to experiment with different pickup heuristics to find one that is more compatible with the training of the route-finding model.

Finally, as described in Sec. \ref{Algorithm}, our algorithm uses two different pickup heuristics. We use a continuous volume pickup heuristic during the route-finding phase. We then forget these pickup decisions before going into the pickup-finding phase where we decide the final pickups with a discrete heuristic. The reasons for this approach are described at the beginning of Sec. \ref{Algorithm}. It is likely that the discrepancy between the final pickup decisions the pickup decisions made during route-finding results in some loss of efficiency. This issue can be addressed in future work with an approach that makes the routing decisions and final pickup decisions at the same time, so that we do not have to go back and make new pickup decisions once the full routes are known.

As always in ML, there is still plenty of exploration that can be done on what hyperparameters give the best performance. These include subenvironment parameters such as the number of nodes and trucks in each subenvironment, model parameters like the encoding dimension and the number of attention heads in each layer, training parameters like the start and minimum learning rate, and many others. The hyperparameters chosen for the results presented here were determined via a combination of Ray Tune \cite{liaw2018tune} (an open-source hyperparameter tuning software) and trial-and-error. However, due to the vastness of the space of hyperparameter combinations, there surely exist configurations which lead to better results. Future work may include further hyperparameter tuning on the model presented in this paper.

\section{Acknowledgements}

The authors thank Natansh Mathur, Victor Putz, and El Amine Cherrat for insightful discussions and assistance on the software side. We also thank Sean Weinberg and Fabio Sanches for initiating the study of this algorithm and for implementing its first version in software.

The authors do disclose the following financial interests. J.L. and R.C. were employed by QC Ware during all or portions of this research. R.C. is a co-founder of QC Ware Corp., J.L. and R.C. hold QC Ware stock or stock options. T.I., Su.T., Sa.T., and A.A. were employed by Aisin Group. Aisin Group is a customer of QC Ware; Aisin Group provided funding for the research presented in this manuscript.

\bibliographystyle{IEEEtran}
\bibliography{main.bib}

\end{document}